\documentclass[conference]{IEEEtran}
\IEEEoverridecommandlockouts

\usepackage{graphicx}
\usepackage{subcaption}
\usepackage{cite}
\usepackage{amsmath,amssymb,amsfonts}
\usepackage{algorithmic}
\usepackage{graphicx}
\usepackage{textcomp}
\usepackage{xcolor}
\usepackage{geometry}
\usepackage{subfig}
\usepackage{multirow}
\def\BibTeX{{\rm B\kern-.05em{\sc i\kern-.025em b}\kern-.08em
    T\kern-.1667em\lower.7ex\hbox{E}\kern-.125emX}}
\begin{document}

\title{Few-Shot Video Recognition via Hierarchical Metric Learning}

\author{
    \IEEEauthorblockN{
        Jiaxin Zhang,
        Haoran Gao,
        Xizhan Gao*,
        Zihao Dong,
        Tingwei Wang,
        Sijie Niu    }
    \IEEEauthorblockA{
        School of Information Science and Engineering, University of Jinan,
        Shandong, China    }
%
%
    \IEEEauthorblockA{
        Corresponding Author: Xizhan Gao. Email: ise\_gaoxz@ujn.edu.cn    }

}

\maketitle

\begin{abstract}
Few-shot action recognition (FSAR) aims to recognize unseen action categories with only a small number of annotated video samples. Recent works typically apply single-prototype supervision at the network output and fail to sufficiently exploit rich cross-frame global spatial information in videos. Even existing multi-level metric schemes only impose parallel prototype constraints on intermediate layers, without progressive supervision along the full feature pipeline, which results in limited generalization ability of the learned class prototypes. Inspired by this, we present a novel method, hierarchical metric learning for few-shot action recognition (HML-FSAR). First, a spatial-enhanced module is developed to capture cross-frame global spatial representations. Combined with temporal MHA, heterogeneous alignment, spatial-temporal feature fusion and dictionary learning modules, it constructs the complete feature processing pipeline. Second, a hierarchical metric learning (HML) strategy is embedded into HML-FSAR. Composed of center metric, alignment metric, contrastive metric, dictionary metric and prototype metric, HML imposes progressive multi-stage complementary constraints from frame-level representations to final class prototypes, so as to jointly optimize feature compactness, heterogeneous spatial-temporal alignment, inter-class discriminability and anti‑noise robustness. The proposed HML-FSAR method is validated on five widely-used FSAR datasets, and experimental results fully demonstrate its effectiveness.
\end{abstract}

\vspace{\baselineskip}


\section{Introduction}

Action recognition aims to automatically identify human actions from video sequences. It has attracted extensive attention due to its wide applications in intelligent surveillance, human--computer interaction, autonomous driving, video retrieval, and healthcare~\cite{simonyan2014twostream, carreira2017Kin}. Compared with image recognition, video action recognition requires models to simultaneously understand spatial appearance and temporal dynamics, making it considerably more challenging. Although recent deep learning methods have achieved remarkable success with large-scale annotated datasets~\cite{carreira2017Kin, bertasius2021timesformer, feichtenhofer2022mvitv2}, collecting sufficient labeled videos is expensive and time-consuming. Therefore, few-shot action recognition (FSAR) has become an important research topic~\cite{Cao2020OTAM, trx2021}.

\begin{figure}[t]
\centering
\includegraphics[width=0.47\textwidth]{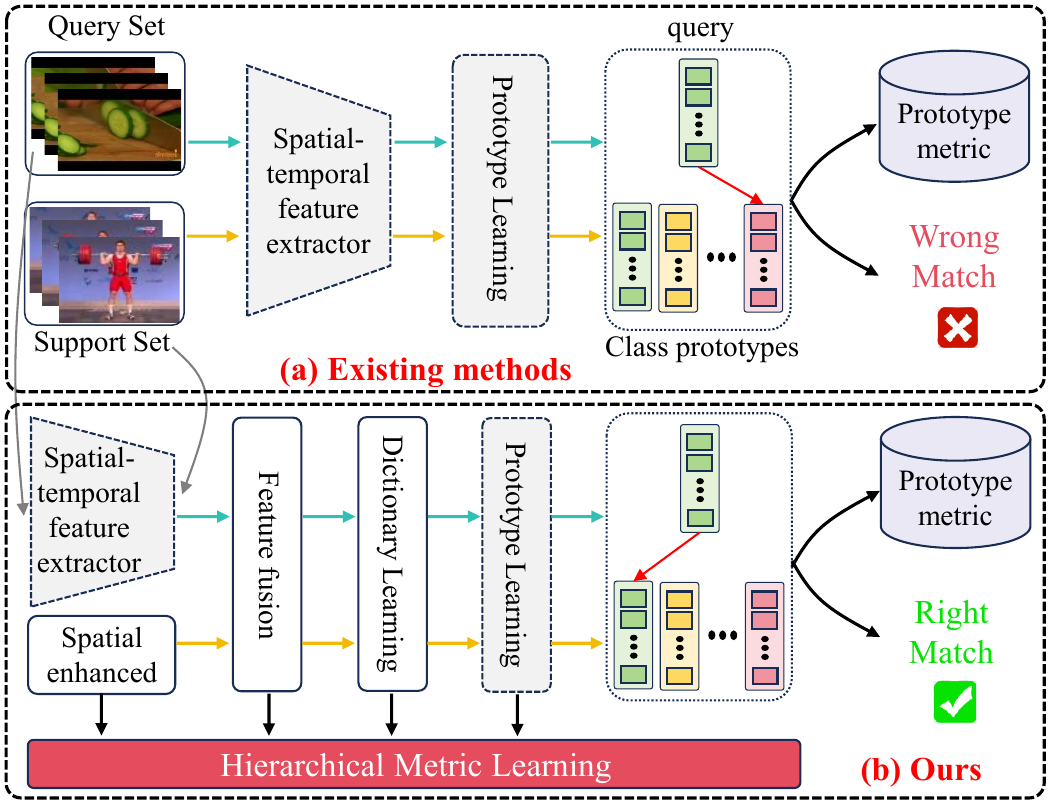}
\caption{Comparison between existing FSAR methods and our HML-FSAR. (a) Existing methods employ only a single prototype metric at the top layer of the network and fail to fully exploit the rich spatial information contained in videos, resulting in poor separability between class prototypes and further leading to misclassification. (b) In contrast, our method introduces a spatial‑enhanced branch to boost the discriminative ability of spatial‑temporal features, and leverages hierarchical metric learning to optimize the prototype feature learning process in multiple stages for obtaining class prototypes with stronger separability, thus achieving more accurate FSAR.}
\label{fig:motivation}
\end{figure}

FSAR mostly adopts the metric‑based meta‑learning paradigm. Through episodic training, videos from the support set and query set are mapped into a unified feature embedding space, and predictions for novel‑category samples are made according to feature similarity (see Fig. \ref{fig:motivation} (a)). Within this framework, the discriminability and generalization ability of class prototypes play a critical role in recognition performance. To obtain prototype features with superior discriminability and generalization performance, existing studies are mainly conducted along two research lines: spatial‑temporal feature learning and metric learning. Spatial‑temporal feature learning methods focus on designing high‑performance spatial‑temporal feature extraction networks, such as TARN \cite{Mina2019TARN}, ProtoGAN \cite{Dwivedi2019ProtoGANTF}, GgHM \cite{xing2023iccv-boosting}, TSAM \cite{li2025aaai-frame}, SFAR \cite{liu2025storyboardguided} and DiST \cite{Qu2026SpatioTemporalDK}. They cover diverse implementations built upon CNN, Transformer, and pre‑trained vision‑language models. Although these methods have achieved promising progress, most of them mine temporal information upon frame‑level local spatial features, or adopt two‑branch networks to learn spatial and temporal features separately. The spatial representations they capture are restricted to local information within individual frames and ignore the benefits of cross‑frame global spatial features for action recognition.

In contrast, metric learning based methods tend to devise diverse metric losses to regularize the learning of spatial-temporal features, such as OTAM \cite{Cao2020OTAM}, MbFSL \cite{careaga2019metricbasedfewshotlearningvideo}, PEAG \cite{Chen2022MLML}, HCL \cite{Zheng2022FSARHML}, and MML‑FSAR \cite{Zheng2026MSMLCLIP}. However, some of these methods \cite{Cao2020OTAM, careaga2019metricbasedfewshotlearningvideo} adopt only a single prototype metric at the network output to guide model parameter updates. Since supervision signals are imposed merely on the final output, intermediate‑layer features can only be updated indirectly via back‑propagation. Such a paradigm tends to drive the network to overemphasize the discriminability of high‑level representations, while early intermediate features lack explicit constraints. Consequently, high‑level features acquire certain discriminative power yet suffer from limited generalization performance. To tackle this issue, several studies have introduced multi‑level metric learning \cite{Chen2022MLML, Zheng2022FSARHML, Zheng2026MSMLCLIP}, which relocates prototype losses to multiple intermediate layers of the network. Nevertheless, such schemes are essentially variants of mutually‑independent multi‑layer prototype metrics that only apply parallel prototype constraints to outputs at different layers. They not only introduce extra computational overhead but also lack progressive relationships among different layers. Hence they cannot fundamentally optimize the evolution of the feature pipeline and merely supplement high‑level representations with middle‑ and low‑level features, which hardly improves the generalization ability of prototypes. We argue that ideal class prototype features should possess both discriminability and generalization capacity. Therefore, supervision should not be limited to the final network output. Instead, explicit progressive constraints should be imposed on the entire feature pipeline from frame‑level to video‑level early stage feature learning.

To address the above limitations, a novel end‑to‑end model named hierarchical metric learning for few-shot action recognition (HML‑FSAR) is proposed. On the one hand, a spatial‑enhanced (SE) module is designed to exploit cross‑frame global spatial representations, compensating for the deficiency of frame‑local spatial representations. This module, together with temporal MHA (TMHA), heterogeneous alignment (HA), spatial‑temporal feature fusion (STFF) and dictionary learning (DL) modules, forms the network architecture of HML‑FSAR. On the other hand, the hierarchical metric learning (HML) strategy is embedded in HML‑FSAR. It consists of center metric, alignment metric, contrastive metric, dictionary metric, and prototype metric, which progressively supervise feature learning along the whole feature propagation pipeline from frame‑level representations to final class prototypes. Specifically, the center metric constrains intra‑video frame‑level feature compactness, the alignment metric eliminates heterogeneity gaps between spatial‑enhanced and temporal features, the contrastive metric improves inter‑class discriminability of fused spatial‑temporal representations, and the dictionary metric suppresses feature noise and mitigates prototype drift under limited labeled samples. Extensive experiments demonstrate the superiority of our proposed method.

The main contributions of this paper are summarized as follows:

\begin{itemize}
\item A spatial‑enhanced module is proposed to mine cross‑frame global spatial representations for videos. It alleviates the limitation that existing approaches only exploit local spatial information within individual frames, and enriches spatial representations for few‑shot action recognition.

\item A novel hierarchical metric learning (HML) strategy is proposed. Unlike existing parallel multi‑layer prototype metrics, it applies progressive complementary constraints (center, alignment, contrastive, and dictionary metrics) across the full feature pipeline to improve feature compactness, alignment, discriminability and robustness.

\item An end‑to‑end model termed HML‑FSAR is constructed, which integrates the above‑mentioned HML strategy, along with our newly‑designed SE, TMHA, HA, STFF and DL modules. Extensive experiments demonstrate its overall effectiveness on five standard benchmarks.
\end{itemize}

The rest of this paper is organized as follows. Section \ref{sec:2} reviews related work. Section \ref{sec:3} elaborates the details of our proposed HML-FSAR method. Section \ref{sec:4} presents experimental results and analysis. Finally, Section \ref{sec:5} concludes this work.

\section{Related Work} \label{sec:2}

\subsection{Few-shot Image Classification}
Few‑shot image classification aims to recognize novel categories with only a small number of labeled images. Existing few‑shot image classification methods can be divided into three categories: data augmentation based methods, meta‑learning based methods, and metric learning based methods. Data augmentation based methods generate additional training samples to alleviate the problem of data scarcity. For example, Wu et al. proposed the MPA \cite{Wu2026MPAMP} method, which leverages large language models to generate diverse paraphrased category descriptions and enriches the support set with supplementary semantic cues. Pintelas et al. proposed the AdaptAugment \cite{PINTELAS2024AdaptAugment} framework, which adaptively expands few‑shot training samples by performing targeted image augmentation on the most hard‑to‑recognize samples. Meta-learning based methods focus on learning transferable prior knowledge from meta‑training episodes. MAML \cite{Finn2017MAML} is one of the most representative meta‑learning based methods. It learns a well‑suited parameter initialization, enabling the model to rapidly adapt to new tasks with only a few gradient‑update steps on the support set. To tackle the training‑instability drawback of MAML, Antoniou et al. \cite{antoniou2018how} proposed the MAML++ model, which mitigates instability in MAML’s optimization phase by improving the gradient propagation process. Metric learning based methods map samples into an embedding space and perform recognition according to feature similarity. The prototypical network \cite{snell2017prototypical} serves as one of the most classic baselines in this field. It learns a prototype for each category in the embedding space and obtains category predictions based on the minimum distance between query samples and these prototypes. Cheng et al. \cite{Cheng2025LPN} proposed the LPN method, which leverages the complementarity between visual and language modalities through two parallel branches to realize few‑shot image classification.

Although these approaches achieve promising performance on image tasks, they cannot be directly applied to video‑based action recognition, since videos contain complex temporal dynamics and cross‑frame spatial dependencies.

\subsection{Few-shot Action Recognition}
Different from the aforementioned few‑shot image classification methods, few‑shot action recognition deals with more complex 3‑dimensional video data instead of 2‑dimensional images. Most existing FSAR methods adopt the metric based meta‑learning paradigm. They first learn spatial‑temporal feature vectors from action videos and project them into an embedding space. Afterwards, prototype learning is conducted within this embedding space to construct class prototypes for classification. These methods can be further divided into spatial‑temporal feature learning methods and metric learning based methods. Spatial-temporal feature learning methods focus on designing high-performance spatial-temporal feature extraction networks. For instance, TARN \cite{Mina2019TARN} employs the C3D network to extract spatial features and utilizes bidirectional GRU to capture temporal features for obtaining videos' spatial‑temporal representations.  TSAM \cite{li2025aaai-frame} introduces a sequence‑aware adapter to fuse spatial information and temporal dynamics into feature embeddings, aiming to produce more discriminative feature representations for video actions. DiST \cite{Qu2026SpatioTemporalDK} aligns frame‑level spatial features with spatial descriptive texts and cross‑frame temporal features with temporal descriptive texts, so as to construct feature representations with rich spatial‑temporal semantics. The metric learning based FSAR methods tend to devise diverse metric losses to regularize the learning of spatial-temporal features. For example, to alleviate matching errors caused by video temporal misalignment, OTAM \cite{Cao2020OTAM} introduces a temporal alignment loss at the top of the network to replace the original prototype loss. To address the problem that high‑level features possess strong discriminability yet poor generalization performance, Zheng et al.  \cite{Zheng2026MSMLCLIP} proposed the MML‑FSAR method. It extracts features from multiple stages of CLIP and learns dedicated similarity metrics for features at each level, leveraging the complementary properties of features across different abstraction levels to optimize few‑shot matching.

Although the above methods have achieved promising progress, there is still room for further improvement in spatial‑temporal feature modeling and feature metric learning. Motivated by this observation, this paper proposes the HML‑FSAR method.

\begin{figure*}[t]
\centering
\includegraphics[width=0.95\textwidth]{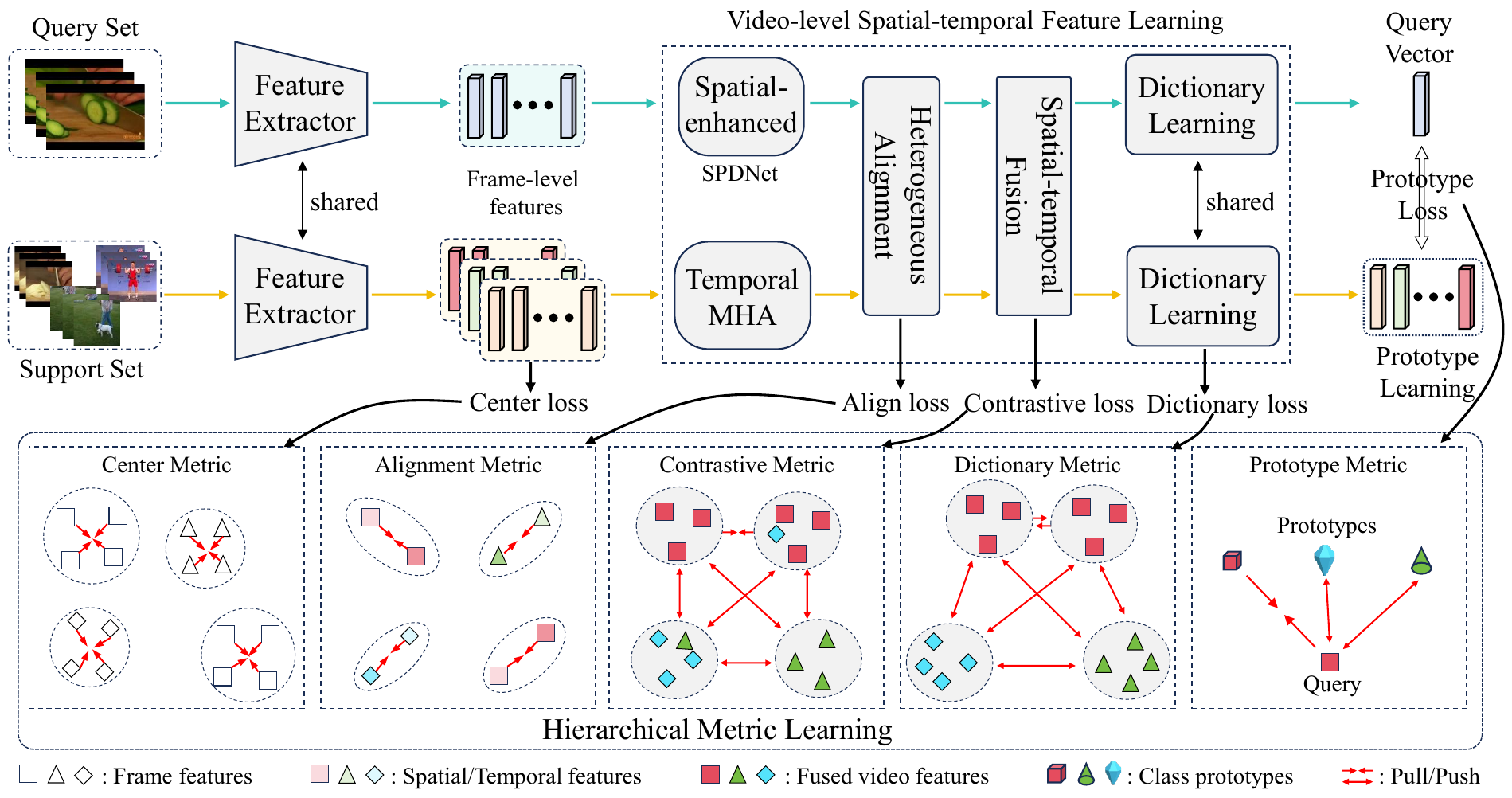}
\caption{Overview of the proposed HML-FSAR method. \emph{Top}: Given input support/query videos, the feature extractor first outputs frame‑level features, and SE and TMHA modules are then adopted in parallel to learn video-level global spatial features and preliminary spatial‑temporal features, respectively. Highly discriminative video‑level features are obtained via HA, STFF and DL modules, followed by prototype learning to generate class prototypes for FSAR. \emph{Bottom}: Hierarchical metric learning, composed of center metric, alignment metric, contrastive metric, dictionary metric and prototype metric, imposes multi‑level metric constraints to progressively refine feature representations throughout the above pipeline.   }
\label{fig:framework}
\end{figure*}

\section{Method} \label{sec:3}
Few-shot action recognition aims to classify video samples from novel action categories using only a few annotated support samples. Following the standard episode learning paradigm, a dataset is divided into a meta-training set $\mathcal{D}_{\text{train}}$ and a meta-testing set $\mathcal{D}_{\text{test}}$ with disjoint action classes, \textit{i.e.}, $\mathcal{D}_{\text{train}} \cap \mathcal{D}_{\text{test}} = \varnothing$. In each training iteration, an $N$-way $K$-shot episode is sampled from $\mathcal{D}_{\text{train}}$, consisting of a support set $\mathcal{S} = \{ (S_i, y_i) \}_{i=1}^{N \cdot K} $ with $K$ labeled videos per class across $N$ categories, and a query set $\mathcal{Q} = \{ Q_j \}_{j=1}^M$ for evaluation, where $S_i = [s_i^1, \cdots, s_i^T]$ denotes the $i$-th support video, which contains $T$ frames; $Q_j = [q_j^1, \cdots, q_j^T]$ represents the $j$-th query video. The objective is to construct a generalizable feature representation space where intra-class samples cluster tightly while inter-class boundaries remain clear and separable.

\subsection{Overall Framework of HML-FSAR}
As shown in Fig. \ref{fig:framework}, to learn more accurate prototype features, we first designed a novel HML-FSAR model, which consists of a feature extractor (FE) module, a spatial-enhanced (SE) module, a temporal MHA (TMHA) module, a heterogeneous alignment (HA) module, a spatial‑temporal feature fusion (STFF) module, and a dictionary learning (DL) module.

More specifically, the FE module is built upon a pre‑trained MAE network to extract frame‑level features from videos. The SE module contains two SPDNet layers \cite{huang2017spdnet} to learn video‑level spatial representations. Stacked with three temporal Transformer blocks, the TMHA module captures preliminary spatial‑temporal (PST) features. The HA module employs two parallel MLP layers to project heterogeneous spatial‑temporal features into a shared feature space. As a two‑layer Transformer block, the STFF module achieves feature fusion through the self‑attention mechanism. The DL module utilizes two successive MLP layers to eliminate interfering components in fused features and enhance the discriminability and generalization performance of features.

The input support/query video is first fed into the FE, SE and TMHA modules to extract spatial‑temporal representations, followed by heterogeneous alignment and self‑attention based feature fusion. After that, the dictionary learning module further refines the fused features for subsequent prototype generation. Nevertheless, relying solely on the prototype metric constraint applied merely to final embeddings will limit the model’s ability to learn well‑separated action prototypes under few-shot settings. To alleviate the above issues, this paper introduces a hierarchical metric learning strategy consisting of the center metric, alignment metric, contrastive metric, dictionary metric and prototype metric, which imposes complementary metric constraints at multiple intermediate stages of feature propagation. Specifically, the center metric is applied to frame‑level features to pull sample frames within the same video close to each other, mitigate intra‑video diversity, and reduce the learning difficulty of subsequent video‑level features. The alignment metric constrains enhanced spatial features and preliminary spatial‑temporal features to eliminate feature discrepancies between them, laying a foundation for discriminative fused features. The contrastive metric acts on fused spatial‑temporal features to guarantee their discriminative power. The dictionary metric imposes constraints on features after contrastive metric supervision to suppress outliers and noise, improve model robustness, and alleviate over‑fitting. Finally, the prototype metric ensures the discriminative capability of class prototypes.

\subsection{Hierarchical Metric Learning}

\subsubsection{Center Metric}
The input support video $S_i$ and query video $Q_j$ are first fed into the FE module to extract frame-level video features: 
\begin{equation}
\begin{aligned}
F_i^s=[f_i^{s,1}, \cdots, f_i^{s,T}] = {\rm MAE}(S_i), \\
F_j^q=[f_j^{q,1}, \cdots, f_j^{q,T}] = {\rm MAE}(Q_j),
\end{aligned}
\label{eq:1}
\end{equation}
where MAE denotes the pre-trained FE module, $f_i^{s,k} \in \mathbb{R}^{d}, (k=1,\cdots, T)$ denotes the frame-level feature, i.e., the $[cls]$ token, of the $k$-th frame from the $i$-th support video, $d$ means the dimension of the feature. $f_j^{q,k} \in \mathbb{R}^{d}$ has a similar definition. It is worth noting that to avoid redundant descriptions, all subsequent modules and metric constraints adopt identical network parameters and computation pipelines for both support and query samples. In what follows, we do not distinguish between support and query samples, and the subscript $i$ denotes any input video.

Due to temporal redundancy and cluttered backgrounds, features of different frames within the same video fluctuate under the influence of shooting viewpoints, motion amplitudes and background noise, thereby triggering remarkable intra‑video diversity. Without constraints on this issue, the discriminative capability of subsequent video‑level features will be degraded. To address this issue, we adopt the following center loss to enhance the compactness of frame‑level features within each video after obtaining frame‑wise representations, which facilitates the subsequent learning of video‑level features:
\begin{equation}
\begin{aligned}
\mathcal{L}_{\text{center}} = \frac{1}{T} \sum_{k=1}^{T} \left\| f_{i}^k - \mu_i \right\|_2^2,
\end{aligned}
\label{eq:2}
\end{equation}
where normalized $f_i^{k} = \frac{f_{i}^k}{\| f_{i}^k \|_2}$ denotes the $k$-th frame‑level feature of the $i$-th video (either support or query video), and $\mu_i$ denotes the feature center of the $i$-th video, which is formulated as follows:
\begin{equation}
\begin{aligned}
\mu_i = \frac{1}{T} \sum_{k=1}^{T} f_i^{k}.
\end{aligned}
\label{eq:3}
\end{equation}
As illustrated in the bottom‑left of Fig. \ref{fig:framework}, the center metric encourages features from frames within the same video to cluster closely, which indirectly enlarges the feature distance between different videos.

\subsubsection{Alignment Metric}
After obtaining frame‑level features, conventional methods generally adopt temporal Transformers to extract spatial‑temporal features of videos. However, the spatial features yielded in this way are only local representations within individual frames and cannot capture video‑level global spatial characteristics. To address this limitation, we design a parallel SE module to mine cross‑frame video‑level spatial information and generate spatial‑enhanced features, which compensates for the deficiency of the temporal branch in global spatial modeling. Specifically, we adopt the covariance matrix $C_i \in \mathbb{R}^{d \times d}$ of each frame‑level feature to construct video‑level spatial representations:
\begin{equation}
\begin{aligned}
C_i = \frac{1}{T - 1} \sum_{k=1}^{T} (f_{i}^k - \mu_i)(f_{i}^k - \mu_i)^T + \epsilon I_d,
\end{aligned}
\label{eq:4}
\end{equation}
where $\epsilon I_d$ is a perturbation term to ensure that $C_i$ is a strictly symmetric positive definite (SPD) matrix. Because an SPD matrix resides on a Riemannian manifold, a two-layer SPDNet architecture \cite{huang2017spdnet} is employed to project $C_i$ into a compact spatial feature matrix $\bar{C}_i = \text{SPDNet}(C_i) \in \mathbb{R}^{d_2 \times d_2}$, where $d_2 \times d_2$ denotes the spatial embedding dimension.

To obtain PST feature, we introduce a learnable sequence token $[Seq]$ and construct the input sequence of the TMHA module as $\left[Seq, F_i^s \right]$. Then TMHA outputs:
\begin{equation}
\begin{aligned}
V_i = {\rm TMHA}(\left[Seq, F_i^s \right]),
\end{aligned}
\label{eq:5}
\end{equation}
and the PST feature can be obtained as $v_i = (V_i)_0 \in \mathbb{R}^{d_3}$.

Since the enhanced-spatial feature $\bar{C}_i \in \mathbb{R}^{d_2 \times d_2}$ (derived from a Riemannian manifold) and PST feature $v_i \in \mathbb{R}^{d_3}$ (derived from Euclidean space) lie in heterogeneous embedding spaces, they cannot be fused directly. To solve this problem, we design the HA module, which can be regarded as a pair of kernel functions $\phi_s(\cdot)$ and $\phi_t(\cdot)$, to project them into a shared Euclidean space $\mathbb{R}^{d_4}$. That is 
\begin{equation}
\begin{aligned}
\bar{c}_i &= \phi_s(\bar{C}_i)  \in \mathbb{R}^{d_4}, \\
\bar{v}_i &= \phi_t(v_i ) \in \mathbb{R}^{d_4}.
\end{aligned}
\label{eq:6}
\end{equation}
Then, $\bar{c}_i$ and $\bar{v}_i$ are normalized via $\bar{c}_i = \frac{ \bar{c}_i }{\| \bar{c}_i \|_2}$, $\bar{v}_i = \frac{ \bar{v}_i }{\| \bar{v}_i \|_2}$.

To guarantee the alignment performance of the HA module and narrow the representation gap between spatial features $\bar{C}_i$ and PST features $v_i$, we design an alignment metric, which consists of two components: an instance‑level alignment term and a distribution‑level \cite{sun2016RFEDA} alignment term:
\begin{equation}
\begin{aligned}
\mathcal{L}_{\text{align}} = &\frac{1}{B} \sum_{i=1}^{B} (1 - \bar{c}_i^T \bar{v}_{i}) + \frac{\lambda}{4 d_4^2} \| C_{spa} - C_{tem} \|_F^2,
\end{aligned}
\label{eq:7}
\end{equation}
where $B$ denotes the batch size, $C_{spa} = \frac{1}{B - 1} \sum_{i=1}^{B} \bar{c}_i \bar{c}_i^T \in \mathbb{R}^{d_4 \times d_4}$ and $C_{tem} = \frac{1}{B - 1} \sum_{i=1}^{B} \bar{v}_i \bar{v}_i^T \in \mathbb{R}^{d_4 \times d_4}$ represent the mini-batch second-order moment matrices, respectively, and $\lambda$ is a balancing hyperparameter. The first term in Eq. (\ref{eq:7}) is the instance‑level alignment term, which constrains the two types of features corresponding to the same video to be as close as possible in the shared space. The second term is the distribution‑level alignment term, which is used to align the global distributions of spatial and temporal features under the mini‑batch samples.

\subsubsection{Contrastive Metric}
Once aligned in the shared latent space $\mathbb{R}^{d_4}$, the projected enhanced-spatial feature $\bar{c}_i$ and PST feature $\bar{v}_i$ are concatenated with a learnable fusion token $r_{\text{fusion}} \in \mathbb{R}^{d_4}$ to construct a sequence $[r_{\text{fusion}}, \bar{c}_i, \bar{v}_i]$. This sequence is fed into the STFF module to yield the fused spatial‑temporal feature:
\begin{equation}
\begin{aligned}
r_i = {\rm STFF}([r_{\text{fusion}}, \bar{c}_i, \bar{v}_i]) \in \mathbb{R}^{d_4}.
\end{aligned}
\label{eq:8}
\end{equation}

Under the few‑shot action recognition setting, training samples are scarce. The fused features obtained merely through spatial‑temporal feature alignment and fusion are prone to inter‑class feature confusion. To this end, we introduce the contrastive metric to reduce the feature distance between samples of the same action and enlarge the feature distance among samples from different actions, thereby enhancing the discriminative performance of fused representations. The contrastive metric is formulated as:
\begin{equation}
\begin{aligned}
\mathcal{L}_{\text{contrast}} = y_{ij} \|r_i - r_j\|_2^2 + (1 - y_{ij}) \max(0, h - \|r_i - r_j\|_2^2),
\end{aligned}
\label{eq:9}
\end{equation}
where $r_j \in \mathbb{R}^{d_4}$ denotes the fused feature of the $j$-th video $j$, $y_{ij} \in \{0, 1\}$ is a binary indicator denoting whether video $i$ and video $j$ belong to the same category ($y_{ij}=1$ if $y_i=y_j$ and $0$ otherwise), and $h > 0$ represents a predefined distance threshold.

\subsubsection{Dictionary Metric}
In FSAR scenarios, each action class contains only a small number of samples. Meanwhile, affected by background interference, viewpoint variations and other factors, the videos' fused features often contain redundant noise. If class prototypes are directly generated by averaging these limited samples, estimation bias is very likely to be introduced, which causes prototype drift and ultimately degrades the accuracy of metric matching. To address this issue, this section designs a DL module together with a dictionary metric constraint. Within this module, sample reconstruction is employed to filter out noisy feature components, so as to provide high‑quality features with less noise and stronger robustness for subsequent prototype learning.

More specifically, for the fused video-level feature $r_i$, an analysis dictionary matrix $P \in \mathbb{R}^{D \times d_4}$ and a synthesis dictionary matrix $U \in \mathbb{R}^{d_4 \times D}$ (parameterized as bias-free linear layers, where $D$ denotes number of dictionary atoms) are used to compute a dictionary code $\alpha_i \in \mathbb{R}^D$ and a purified reconstructed feature $\hat{r}_i \in \mathbb{R}^{d_4}$:
\begin{equation}
\begin{aligned}
\alpha_i &= \text{ReLU}(P r_i) \in \mathbb{R}^D, \\
\hat{r}_i &= U \alpha_i = U \cdot \text{ReLU}(P r_i) \in \mathbb{R}^{d_4},
\end{aligned}
\label{eq:10}
\end{equation}
where $\text{ReLU}(\cdot)$ denotes the rectified linear unit activation function.

To guarantee reconstruction quality and enhance the discriminative capability of dictionary codes, the dictionary metric constraint is formulated as follows:
\begin{equation}
\begin{aligned}
\mathcal{L}_{\text{dict}} = \mathcal{L}_{\text{recon}} + \lambda_1 G_1 + \lambda_2 G_2,
\end{aligned}
\label{eq:11}
\end{equation}
where $\lambda_1, \lambda_2$ are hyperparameter weights balancing different constraints. The three loss terms are defined as follows.

\textbf{Reconstruction Loss $\mathcal{L}_{\text{recon}}$}:
    \begin{equation}
    \mathcal{L}_{\text{recon}} = \frac{1}{B d_4} \sum_{i=1}^{B} \|\hat{r}_i - \text{sg}[r_i]\|_2^2,
    \end{equation}
where $\text{sg}[\cdot]$ denotes the stop-gradient operation to stabilize dictionary parameter learning, ensuring that gradient only goes through $\hat{r}_i$ during back‑propagation.

\textbf{Discriminative Code Constraint $G_1$}:
    \begin{equation}
    \begin{aligned}
    G_1 = &\frac{1}{|\mathcal{C}|} \sum_{c \in \mathcal{C}} \left( \frac{1}{|\Omega_c|} \sum_{i \in \Omega_c} \|\alpha_i - \mu_c\|_2^2 \right) \\
    &- \frac{0.1}{|\mathcal{C}|^2} \sum_{c, c' \in \mathcal{C}} \|\mu_c - \mu_{c'}\|_2,
    \end{aligned}
    \end{equation}
where $\mathcal{C}$ denotes the set of action categories in the episode, $|\mathcal{C}| = N$ is the total number of categories, $\Omega_c$ is the set of sample indices belonging to class $c$, $|\Omega_c|$ is the number of samples in class $c$, and $\mu_c = \frac{1}{|\Omega_c|} \sum_{i \in \Omega_c} \alpha_i \in \mathbb{R}^D$ represents the mean dictionary code for class $c$.

\textbf{Dictionary Orthogonality Constraint $G_2$}:
    \begin{equation}
    G_2 = \|P P^T - I_D\|_F^2,
    \end{equation}
where $I_D \in \mathbb{R}^{D \times D}$ is the identity matrix, and $\|\cdot\|_F$ denotes the Frobenius norm.

\subsubsection{Prototype Metric}
Finally, the prototype metric is adopted to generate class prototypes and perform few‑shot video action classification. Specifically, using the purified feature representations $\hat{r}_i$, class prototypes $g_c \in \mathbb{R}^{d_4}$ for the $c$-th class are generated by averaging support embeddings in the support subset $\mathcal{S}_c$:
\begin{equation}
g_c = \frac{1}{K} \sum_{i \in \mathcal{S}_c} \hat{r}_i,
\end{equation}
where \(\mathcal{S}_{c}\) denotes the subset of support samples belonging to class $c$, and $K$ denotes the number of support videos per class.

For the $j$-th query video $Q_j \in \mathcal{Q}$ with ground‑truth class label $y_{q_j}$, the video is fed into the proposed feature extraction and dictionary learning pipeline to obtain the purified reconstructed feature $q_j =\hat{r}_j^q \in\mathbb{R}^{d_4}$. Then the prototype metric is formulated as:
\begin{equation}
\begin{aligned}
\mathcal{L}_{\text{prot}} = - \frac{1}{M} \sum_{j=1}^M \Bigg[ &\tau \cdot \cos(q_j, g_{y_{q_j}}) - \\
&\log \sum_{c=1}^N \exp(\tau \cdot \cos(q_j, g_c)) \Bigg],
\end{aligned}
\label{eq:16}
\end{equation}
where $M$ denotes the total number of query samples in the episode and $\tau > 0$ is a learnable temperature scaling parameter.

\subsection{Network Optimization}
The proposed HML-FSAR method is trained in an end‑to‑end manner. The overall multi‑objective loss function $\mathcal{L}_{\text{total}}$ combines the main prototype loss with auxiliary losses, including center, alignment, contrastive, and dictionary losses:
\begin{equation}
\begin{aligned}
\mathcal{L}_{\text{total}} = \mathcal{L}_{\text{prot}} &+ \beta_1 \mathcal{L}_{\text{center}} + \beta_2 \mathcal{L}_{\text{align}} \\
&+ \beta_3 \mathcal{L}_{\text{contrast}} + \beta_4 \mathcal{L}_{\text{dict}},
\end{aligned}
\label{eq:17}
\end{equation}
where $\beta_1, \beta_2, \beta_3,$ and $\beta_4$ are balancing hyperparameters controlling the relative trade-offs among the auxiliary loss constraints.

\section{Experiments} \label{sec:4}

\subsection{Experimental Settings}
\textbf{Datasets.}
We evaluate the proposed method on five widely used few-shot action recognition benchmarks, including HMDB51 \cite{kuehne2011hmdb}, UCF101 \cite{soomro2012ucf101}, Kinetics \cite{carreira2017Kin}, SSv2-Full and SSv2-Small, both of which are constructed from the Something-Something V2 dataset \cite{goyal2017something}. Among them, HMDB51, UCF101, and Kinetics primarily emphasize scene-level semantic understanding, whereas SSv2 requires stronger temporal modeling capability because the recognition of many actions mainly depends on motion cues rather than object appearance.

For HMDB51 and UCF101, we follow the few-shot split protocol adopted in literature \cite{Cao2020OTAM}, where the numbers of training, validation, and test classes are 31/10/10 and 70/10/21, respectively. For Kinetics, we adopt the subset split strategy following literature \cite{wang2022hybrid}, which divides the classes into 64/12/24 for training, validation, and testing. For SSv2-Full and SSv2-Small, both benchmarks are constructed by selecting 100 classes from the original Something-Something V2 dataset \cite{goyal2017something}, with 64/12/24 classes used for training, validation, and testing, respectively. Compared with SSv2-Small, SSv2-Full contains more than ten times more training videos per class.

\begin{table*}[t]
\centering
\caption{Performance comparison against state‑of‑the‑art methods on five FSAR benchmarks.}
\label{tab:sota_comparison}
\resizebox{\textwidth}{!}{
\begin{tabular}{l|c|cc|cc|cc|cc|cc}
\hline
\multirow{2}{*}{Method} & \multirow{2}{*}{Reference}
& \multicolumn{2}{c|}{HMDB51}
& \multicolumn{2}{c|}{UCF101}
& \multicolumn{2}{c|}{Kinetics}
& \multicolumn{2}{c|}{SSv2-Small}
& \multicolumn{2}{c}{SSv2-Full} \\
\cline{3-12}
& & 1-shot & 5-shot & 1-shot & 5-shot & 1-shot & 5-shot & 1-shot & 5-shot & 1-shot & 5-shot \\
\hline
ProtoNet        \cite{snell2017prototypical}      & NeurIPS'17 & 54.2 & 68.4 & 74.0 & 89.6 & 64.5 & 77.9 & --   & --   & --   & --   \\
CMN             \cite{zhu2018compound}            & ECCV'18    & --   & --   & 57.3 & 76.0 & 34.4 & 43.8 & 36.2 & 48.9 & --   & --   \\
OTAM            \cite{Cao2020OTAM}                 & CVPR'20    & 54.5 & 68.0 & 79.9 & 88.9 & 73.0 & 85.8 & --   & --   & 42.8 & 52.3 \\
AMeFu-Net       \cite{fu2020adaptive}             & MM'20      & 60.2 & 75.5 & 85.1 & 95.5 & 74.1 & 86.8 & --   & --   & --   & --   \\
TARN            \cite{Mina2019TARN}               & ECCV'20    & 45.5 & 60.6 & 66.3 & 83.1 & 63.7 & 82.4 & --   & --   & --   & --   \\
TRX             \cite{perrett2021temporal}        & CVPR'21    & --   & 75.6 & --   & 96.1 & 63.6 & 85.9 & --   & 59.1 & --   & 64.6 \\
SRPN            \cite{wang2021SRPN}               & MM'21      & 61.6 & 76.2 & 86.5 & 95.8 & 75.2 & 87.1 & --   & --   & --   & --   \\
HyRSM           \cite{wang2022hybrid}             & CVPR'22    & 60.3 & 76.0 & 83.9 & 94.7 & 73.7 & 86.1 & 40.6 & 56.1 & 54.3 & 69.0 \\
MTFAN           \cite{wu2022multi}                & CVPR'22    & 59.0 & 74.6 & 84.8 & 95.1 & 74.6 & 87.4 & --   & --   & 45.8 & 60.4 \\
TA2N+Sampler    \cite{li2022ta2n}                 & AAAI'22    & 59.9 & 73.5 & 83.5 & 96.0 & 73.6 & 86.2 & --   & --   & 47.1 & 61.6 \\
MoLo            \cite{wang2023molo}               & CVPR'23    & 60.8 & 77.4 & 86.0 & 95.5 & 74.0 & 85.6 & 41.9 & 56.2 & 55.0 & 69.6 \\
MGCSM           \cite{Yu2023MSGCSM}               & MM'23      & 61.3 & 79.3 & 86.5 & 97.1 & 74.2 & 88.2 & --   & --   & --   & --   \\
SA-CT           \cite{zhang2023importance}        & MM'23      & 60.4 & 78.3 & 85.4 & 96.4 & 71.9 & 87.1 & --   & --   & 48.9 & 69.1 \\
CLIP-FSAR       \cite{wang2024clip}               & IJCV'24    & 69.2 & 80.3 & 91.3 & 97.0 & 87.6 & 91.9 & 51.5 & 57.1 & 58.1 & 62.8 \\
CCLN            \cite{wang2024cross}              & TIP'24     & 65.1 & 78.8 & 86.9 & 96.1 & 75.8 & 87.5 & 46.0 & 61.3 & --   & --   \\
MVP-Shot        \cite{qu2025mvp}                  & TMM'25     & 72.5 & 82.5 & 92.2 & 97.6 & 90.0 & 93.2 & 51.2 & 57.0 & 59.9 & 64.1 \\
TEAM            \cite{liu2025team}                & CVPR'25    & 70.9 & 85.5 & 94.5 & 98.8 & 83.3 & 92.9 & 47.2 & 63.1 & --   & --   \\
\hline
\textbf{Ours}   & -- & \textbf{71.2} & \textbf{86.1} & \textbf{93.5} & \textbf{99.2} & \textbf{86.0} & \textbf{96.5} & \textbf{52.0} & \textbf{66.7} & \textbf{58.3} & \textbf{73.2} \\
\hline
\end{tabular}
}
\end{table*}

\textbf{Implementation Details.}
All experiments are conducted on an NVIDIA RTX 3090 GPU. During training, the model is trained for 10,000 episodes on HMDB51, UCF101, and Kinetics, while 60,000 episodes are used for SSv2-Full and SSv2-Small because these datasets require stronger temporal modeling. The learning rate is uniformly set to $1\times10^{-5}$ for all experiments. During testing, 10,000 episodes are sampled on every dataset to ensure stable and reliable evaluation.

\subsection{Comparison with Various Methods}
To verify the effectiveness of the proposed method, we compare its performance with state‑of‑the‑art approaches on five datasets, and the results are reported in Table \ref{tab:sota_comparison}.
As can be seen from the table, the proposed method achieves superior performance under the vast majority of evaluation settings. Notably, consistent and prominent performance improvements are obtained under the 5‑shot setting on all five datasets, which fully demonstrates the effectiveness of our approach for few‑shot action recognition. Specifically, our method achieves $71.2\%/86.1\%$ on HMDB51, $93.5\%/99.2\%$ on UCF101, and $86.0\%/96.5\%$ on Kinetics under the 1-shot/5-shot settings, respectively, outperforming previous state-of-the-art methods in most cases. These results indicate that the proposed method is capable of learning discriminative action representations and exhibits strong generalization ability across datasets with different scales and visual characteristics.

On the more temporally challenging SSv2 benchmarks, our method also achieves the best overall performance, obtaining $52.0\%/66.7\%$ on SSv2-Small and $58.3\%/73.2\%$ on SSv2-Full under the 1-shot/5-shot settings. Compared with the previous state-of-the-art method TEAM, our method improves the performance by $4.8\%$ and $3.6\%$ on SSv2-Small. Compared with MVP-Shot, our method improves the accuracy by $9.1\%$ under 5-shot settings on SSv2-Full dataset, demonstrating its superior capability in modeling fine-grained temporal dynamics. The performance gains mainly stem from our proposed hierarchical metric learning strategy. It imposes multi‑layer metric constraints on video features to optimize intra‑class compactness and inter‑class separability in the feature space. Faced with limited training samples, these hierarchical constraints can guide the network to learn more discriminative video representations and effectively improve the generalization ability of the model, thus achieving excellent performance for few‑shot action recognition.

In addition, compared with methods based on the pre‑trained vision‑language model CLIP (CLIP‑FSAR, MVP‑Shot), our method still achieves performance advantages. Under the 5‑shot setting, the proposed model obtains higher recognition accuracy across all five datasets; under the 1‑shot setting, our method is also highly competitive. This result demonstrates that for the few‑shot action recognition task, the hierarchical metric learning strategy can mine more discriminative spatial‑temporal features, thus effectively improving the generalization performance of the model and achieving recognition capability comparable to or even better than those cross‑modal image‑text‑based competitors.

\subsection{Ablation Study}
To dissect the working mechanism of each design in our HML-FSAR method, we conduct a set of ablation experiments for quantitative analysis. First, we verify the effectiveness of key components and metric strategies via module ablation and comparisons among different metric schemes. Then, we explore the performance variations caused by different input frame numbers and various few‑shot settings. Finally, we carry out hyperparameter sensitivity experiments to analyze the robustness of the model with respect to key hyperparameters.

\begin{table}[!t]
\centering
\scriptsize
\setlength{\tabcolsep}{4pt} 
\caption{Ablation study on different metric learning strategies.}
\label{tab:ablation_metrics}
\begin{tabular}{c|ccccc|c}
\hline
Setting  & Prototype & Center & Alignment & Contrastive & Dictionary  & Kinetics \\
         & Metric & Metric & Metric & Metric & Metric &    \\
\hline
(a) & \checkmark &            &            &            &            & 83.7 \\
(b) & \checkmark & \checkmark &            &            &            & 83.9 \\
(c) & \checkmark & \checkmark & \checkmark &            &            & 84.1 \\
(d) & \checkmark & \checkmark & \checkmark & \checkmark &            & 85.3 \\
(e) & \checkmark & \checkmark & \checkmark & \checkmark & \checkmark & \textbf{86.0}\\
\hline
\end{tabular}
\end{table}

\begin{table}[!t]
\centering
\caption{Ablation study on different components. S, T, F, and D denote the SE module, TMHA module, HA+STFF module, and DL module, respectively.}
\label{tab:ablation_components}
\begin{tabular}{c|cccc|cc}
\hline
Setting & S & T & F & D & Kinetics & SSv2-small\\
\hline
(a) &  &  &  &  & 80.0 & 43.1\\
(b) & \checkmark &  &  &  & 81.8 & 45.3\\
(c) &  & \checkmark &  &  & 83.6 & 48.9\\
(d) & \checkmark & \checkmark & \checkmark &  & 83.9 & 50.3\\
(e) &  &  &  & \checkmark & 80.7 & 44.2\\
(f) & \checkmark & \checkmark & \checkmark & \checkmark & \textbf{86.0} & \textbf{52.0}\\
\hline
\end{tabular}
\end{table}

\begin{table}[!t]
\centering
\caption{Effect of the number of selected frames.}
\label{tab:ablation_frames}
\begin{tabular}{c|cc|cc}
\hline
\multirow{2}{*}{Selected Frames}
& \multicolumn{2}{c|}{Kinetics}
& \multicolumn{2}{c}{SSv2-small}\\
\cline{2-5}
&1-shot&5-shot&1-shot&5-shot\\
\hline
4&82.8&91.3&47.3&62.7\\
8&85.1&93.7&51.2&64.3\\
16&\textbf{86.0}&\textbf{96.5}&\textbf{52.0}&\textbf{66.7}\\
\hline
\end{tabular}
\end{table}

\subsubsection{\textbf{Influence of Different Metric Learning Strategies}}
To verify the effectiveness of our hierarchical metric learning strategy, we conduct ablation comparisons over different metric schemes while keeping the backbone network unchanged. In this group of experiments, we incrementally add each metric constraint, and the results are reported in Table \ref{tab:ablation_metrics}.

When only the prototype metric is employed, features lack explicit constraints for the internal frame distribution within each video. Performance improves notably after adding the center metric, which demonstrates that the center metric can effectively reduce the dispersion of frame‑wise features inside a single video and promote feature aggregation. On this basis, the alignment metric further narrows the distribution offset between spatial and temporal features and achieves better fusion of spatial‑temporal representations. The introduction of the contrastive metric enhances inter‑class discrimination. Finally, the dictionary metric improves feature robustness under limited training samples, and our full hierarchical multi‑metric scheme achieves the best performance.

These results indicate that each metric constraint plays its own complementary role and is not redundant. The center metric optimizes intra‑sample feature aggregation, the alignment metric calibrates heterogeneous spatio‑temporal features, the contrastive metric enlarges inter‑class margins, and the dictionary metric stabilizes feature representations in few‑shot scenarios. Continuous performance improvements are obtained by stacking these metrics layer‑by‑layer, which verifies the rationality and necessity of the proposed hierarchical multi‑metric method.

\subsubsection{\textbf{Ablation on Key Components}}
Table~\ref{tab:ablation_components} reports the contribution of each proposed module. Without introducing any additional modules (Setting (a)), the baseline achieves $80.0\%$ and $43.1\%$ accuracy on Kinetics and SSv2-small datasets, respectively. Incorporating the SE module (Setting (b)) improves the performance to $81.8\%$ and $45.3\%$, demonstrating that the cross-frame global spatial features can provide complementary discriminative information for local frame‑level spatial features. Using only the TMHA module (Setting (c)) further improves the accuracy to $83.6\%$ on Kinetics and $48.9\%$ on SSv2-small, indicating the importance of temporal dynamics for few-shot action recognition.

When the SE, TMHA, HA and STFF modules are jointly employed (Setting (d)), the accuracy is further improved to $83.9\%$ and $50.3\%$, confirming that spatial and temporal representations provide complementary information. Although the DL module alone (Setting (e)) achieves only limited improvement, incorporating it into the complete method (Setting (f)) further boosts the performance to $86.0\%$ on Kinetics and $52.0\%$ on SSv2-small. These results demonstrate that all proposed modules complement each other and jointly contribute to more discriminative and robust video representations.

\subsubsection{\textbf{Effect of the Number of Selected Frames}}
Table~\ref{tab:ablation_frames} presents the influence of the number of sampled frames on model performance. As the number of sampled frames increases from 4 to 16, the recognition accuracy consistently improves on both Kinetics and SSv2-small. Specifically, on Kinetics, the performance increases from $82.8\%/91.3\%$ to $86.0\%/96.5\%$ under the 1-shot/5-shot settings, while on SSv2-small, the corresponding accuracy improves from $47.3\%/62.7\%$ to $52.0\%/66.7\%$. These results indicate that richer temporal observations enable the proposed method to capture more complete motion patterns and long‑range temporal dependencies. Therefore, a default setting of $16$ sampled frames is adopted in all subsequent experiments.

\subsubsection{\textbf{Effect of Different Few-shot Settings}}
\begin{figure}[!t]
\centering

\begin{subfigure}{0.24\textwidth}
    \centering
    \includegraphics[width=\linewidth]{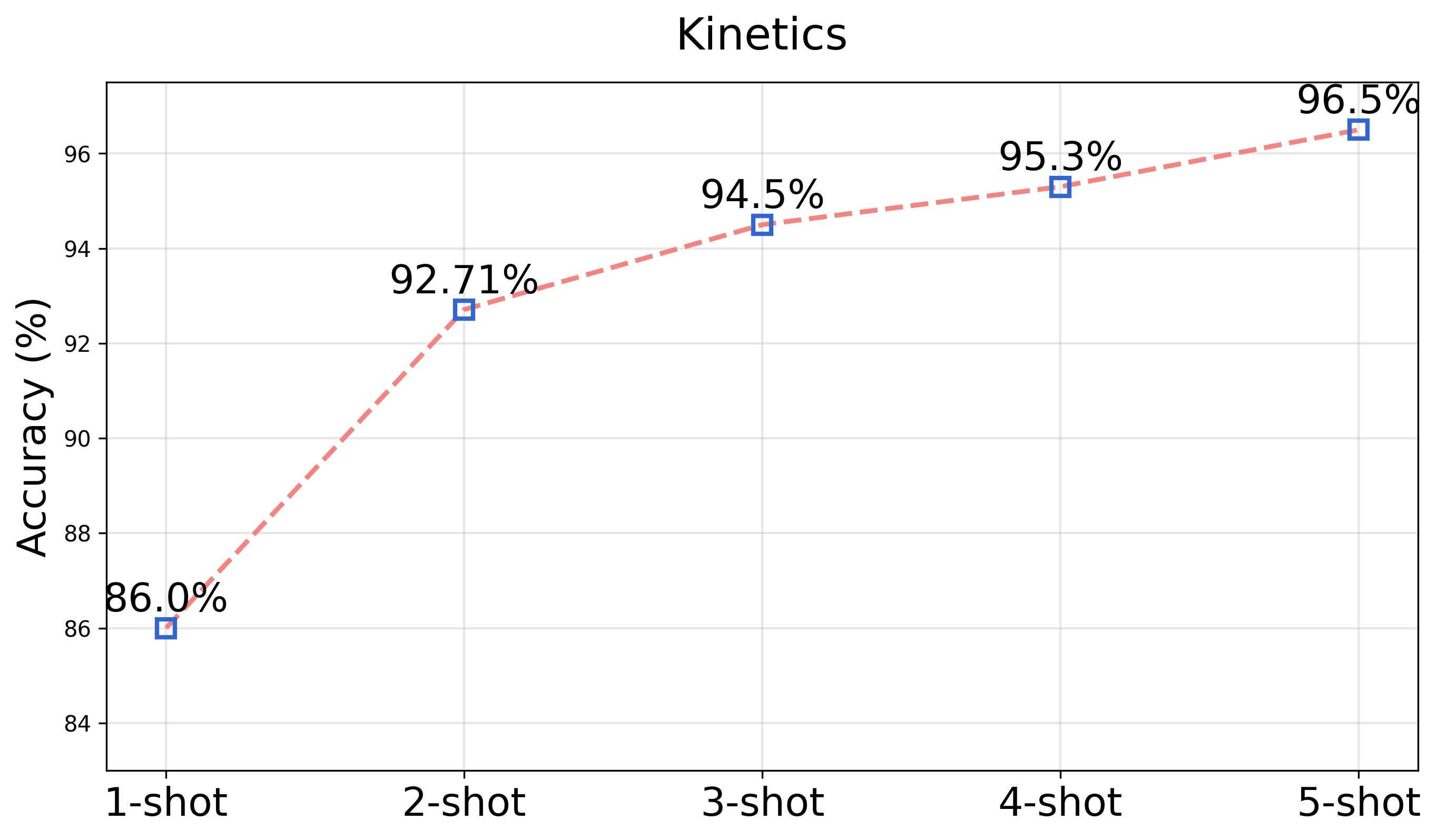}
    \caption{Different shot settings}
    \label{fig:shot}
\end{subfigure}
\hfill
\begin{subfigure}{0.24\textwidth}
    \centering
    \includegraphics[width=\linewidth]{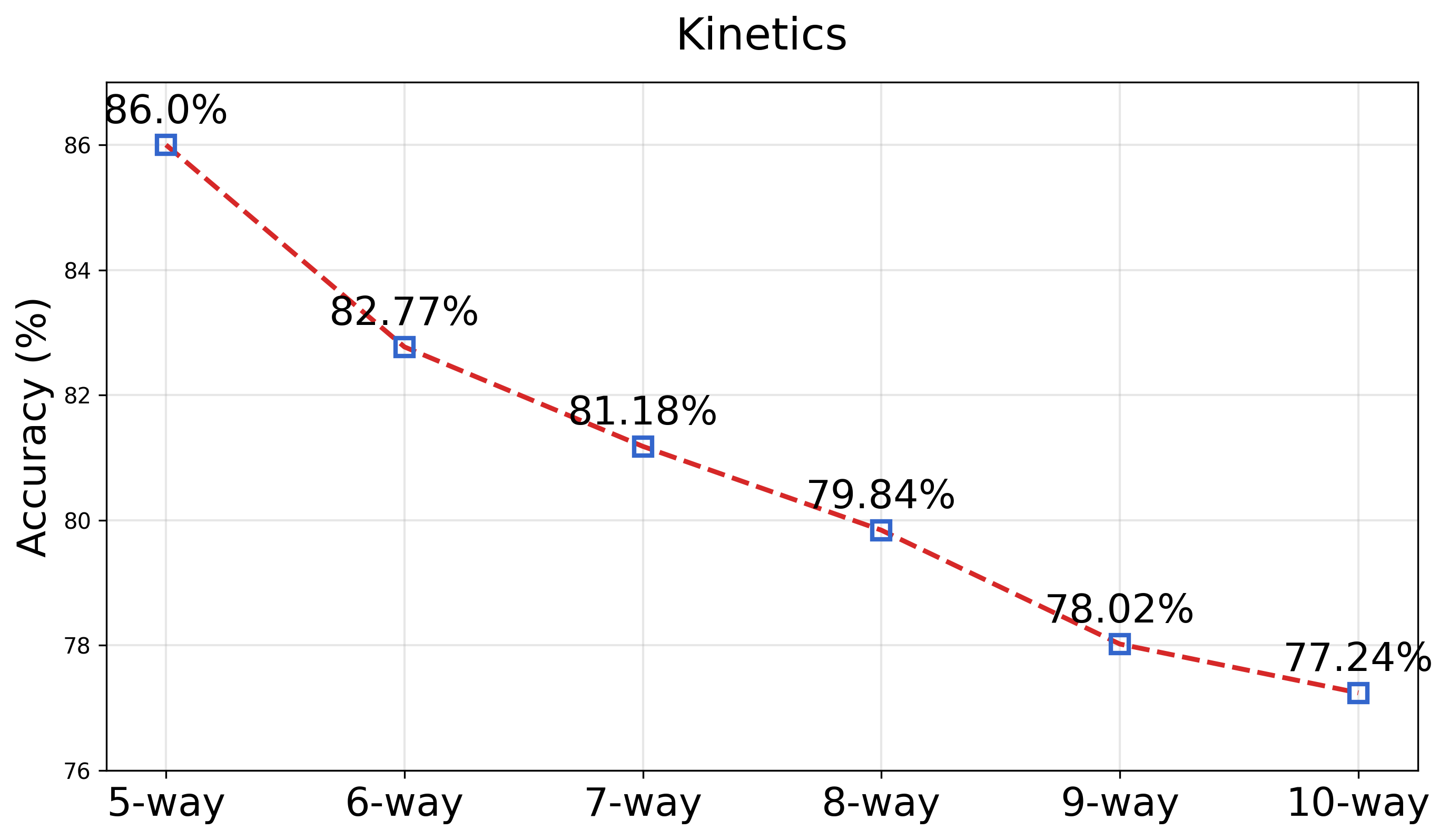}
    \caption{Different way settings}
    \label{fig:way}
\end{subfigure}
\caption{Performance under different few-shot settings on the Kinetics dataset: (a) 5‑way with varying‑shot settings, (b) 1-shot with varying-way settings.}
\label{fig:fewshot}
\end{figure}

To evaluate the robustness of the proposed method under different few-shot task configurations, we further investigate the influence of the number of support samples and the number of action categories on recognition performance. The corresponding results are illustrated in Fig.~\ref{fig:fewshot}, where Fig.~\ref{fig:shot} reports the performance under different shot settings with the number of classes fixed to 5-way, while Fig.~\ref{fig:way} presents the results under different way settings with the number of support samples fixed to 1-shot.

\begin{figure}[!t]
\centering
\begin{subfigure}{0.24\textwidth}
    \centering
    \includegraphics[width=\linewidth,height=3cm]{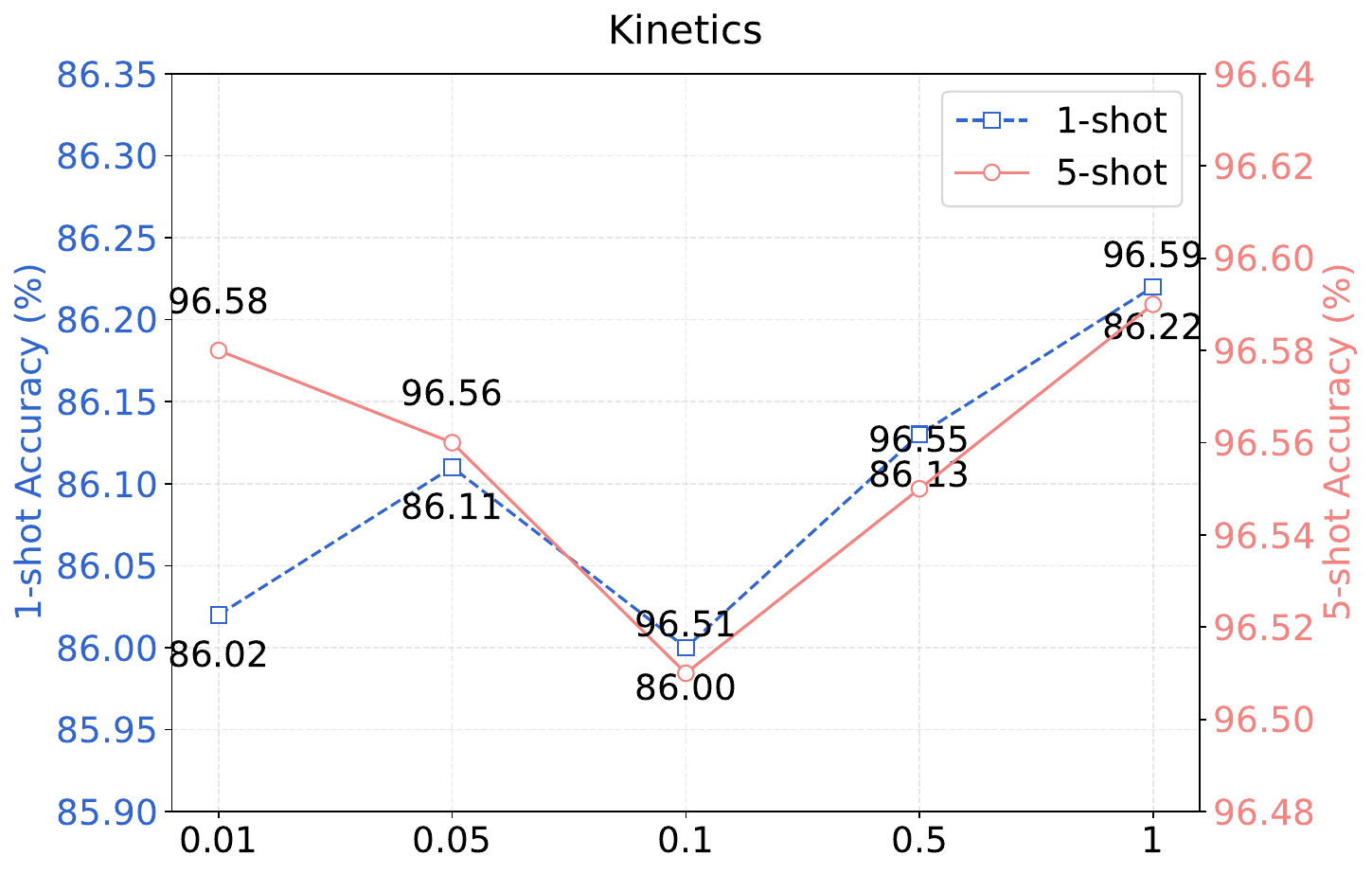}
    \caption{Varying $\lambda$}
    \label{fig:lambda}
\end{subfigure}
\hfill
\begin{subfigure}{0.24\textwidth}
    \centering
    \includegraphics[width=\linewidth,height=3cm]{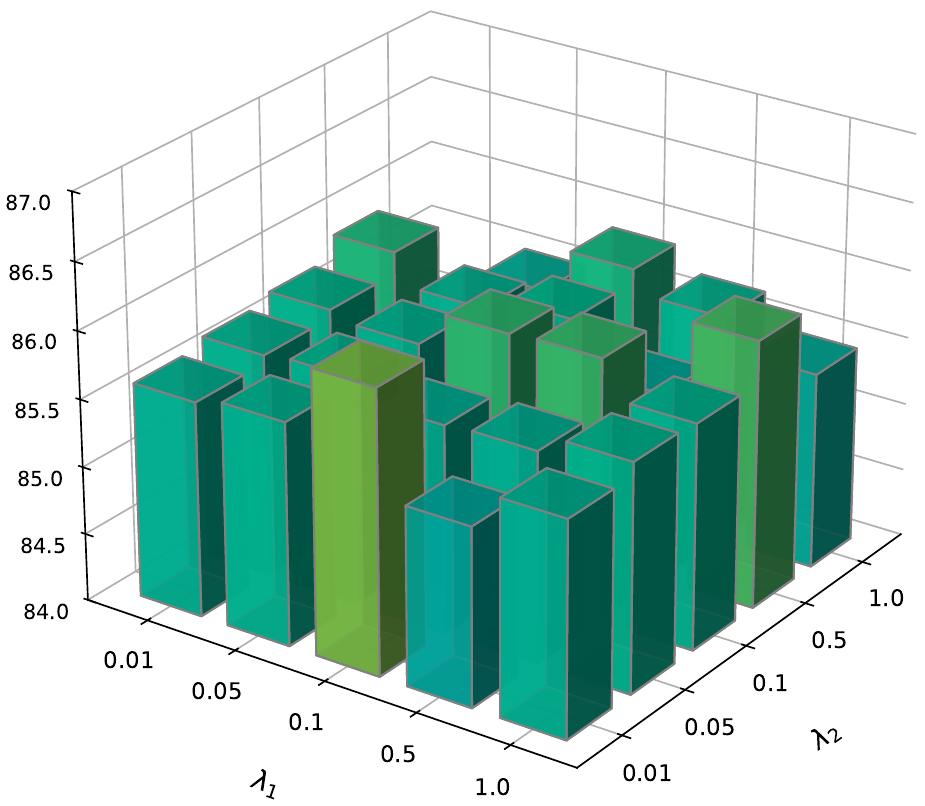}
    \caption{Varying ($\lambda_1$, $\lambda_2$)}
    \label{fig:lambda12}
\end{subfigure}
\caption{Parameter sensitivity analysis on the Kinetics dataset: (a) Varying parameter $\lambda$; (b) Varying combinations of $(\lambda_1, \lambda_2)$.}
\label{fig:param-lambda}
\end{figure}

\begin{figure}[!t]
\centering
\begin{subfigure}{0.24\textwidth}
    \centering
    \includegraphics[width=\linewidth,height=3.2cm]{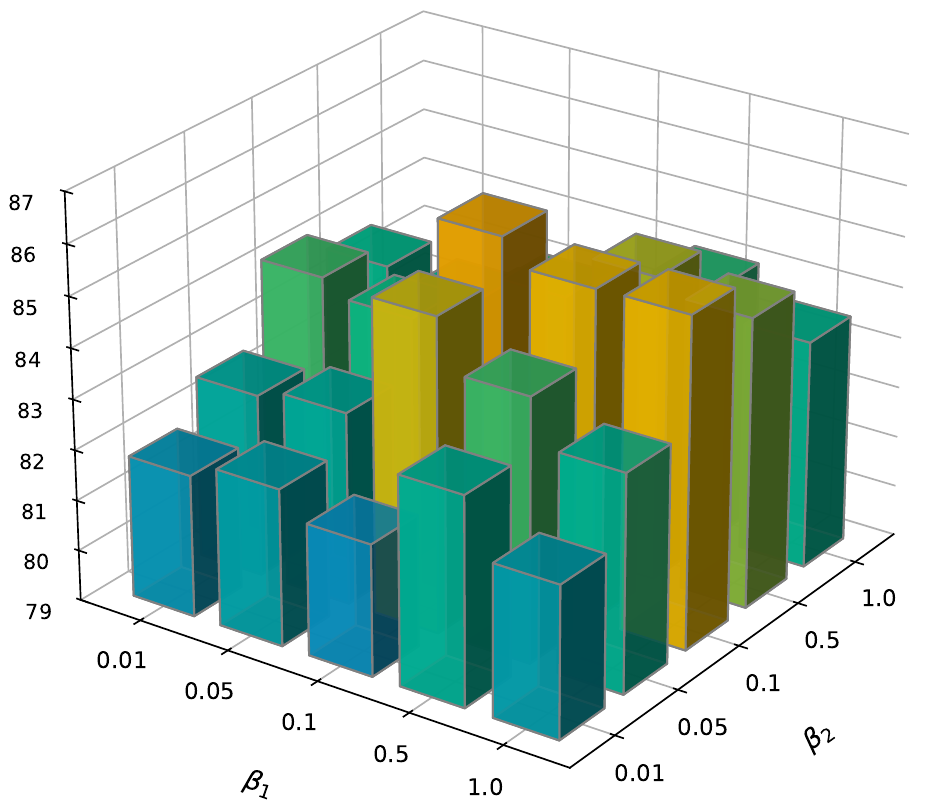}
    \caption{Varying ($\beta_1$, $\beta_2$)}
    \label{fig:beta12}
\end{subfigure}
\hfill
\begin{subfigure}{0.24\textwidth}
    \centering
    \includegraphics[width=\linewidth,height=3.2cm]{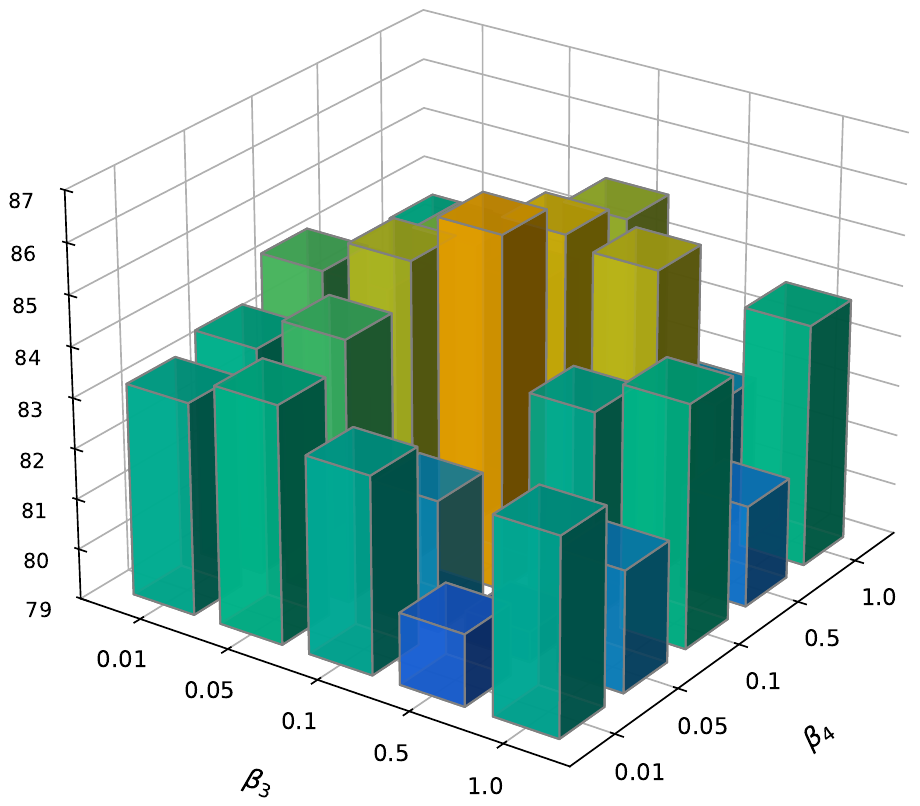}
    \caption{Varying ($\beta_3$, $\beta_4$)}
    \label{fig:beta34}
\end{subfigure}
\caption{Parameter sensitivity analysis on the Kinetics dataset: (a) Varying combinations of $(\beta_1, \beta_2)$; (b) Varying combinations of $(\beta_3, \beta_4)$.}
\label{fig:param-beta}
\end{figure}

As shown in Fig.~\ref{fig:shot}, the recognition accuracy steadily increases from 86.0\% to 96.5\% as the number of support samples grows from 1-shot to 5-shot, demonstrating that additional support samples provide more reliable class prototypes and reduce the adverse effects of intra‑class variations. Conversely, Fig.~\ref{fig:way} shows that the recognition accuracy gradually decreases from 86.0\% to 77.2\% as the number of categories increases from 5-way to 10-way under the fixed 1-shot setting. This performance degradation is expected because distinguishing among more candidate classes becomes increasingly difficult with extremely limited supervision. Nevertheless, the proposed method maintains relatively stable performance across different task settings, demonstrating its strong robustness and generalization ability in few-shot action recognition.

\begin{figure*}[!t]
\centering
\begin{subfigure}{0.19\textwidth}
    \centering
    \includegraphics[width=\linewidth]{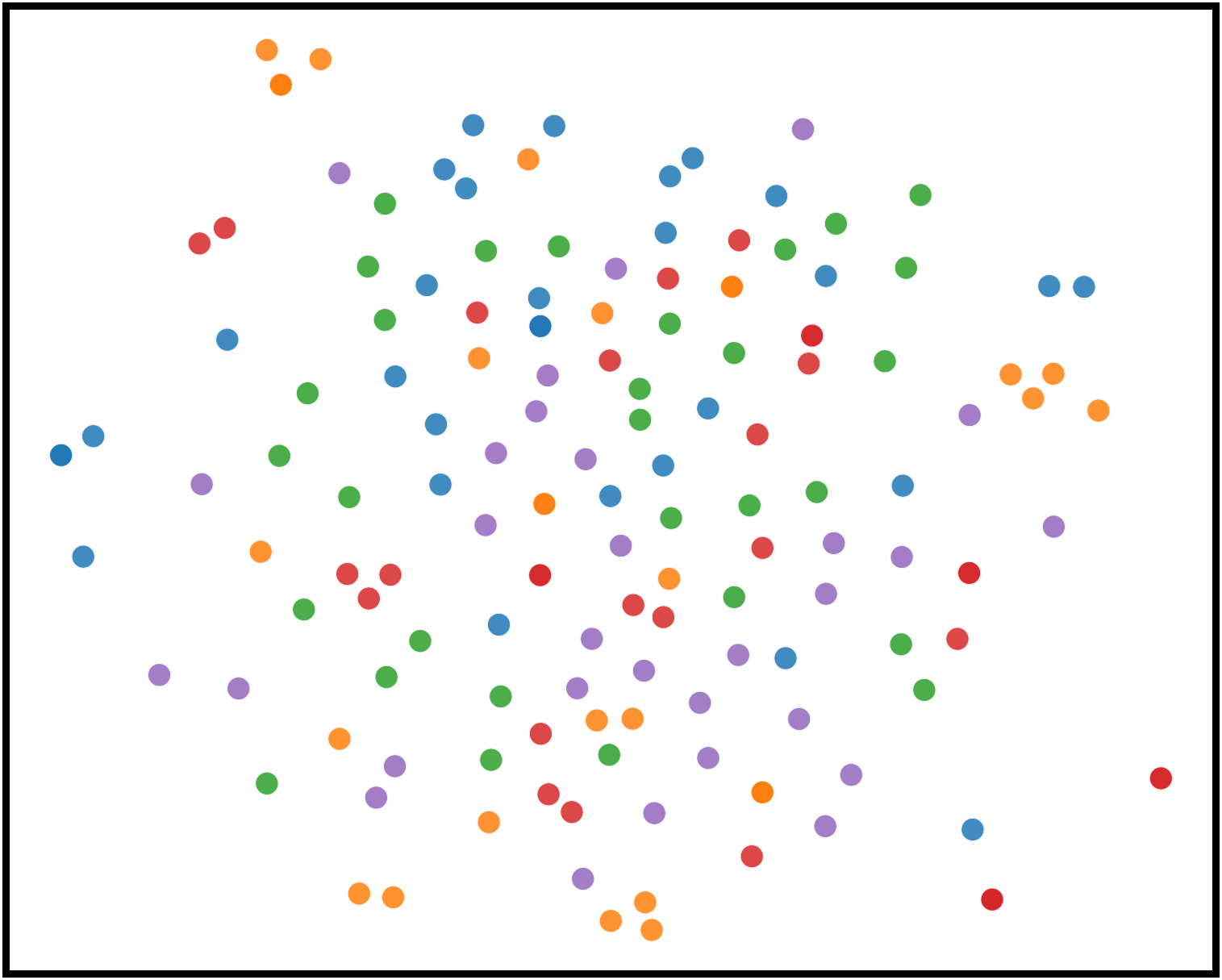}
    \caption{FE module}
    \label{fig:init}
\end{subfigure}
\hfill
\begin{subfigure}{0.19\textwidth}
    \centering
    \includegraphics[width=\linewidth]{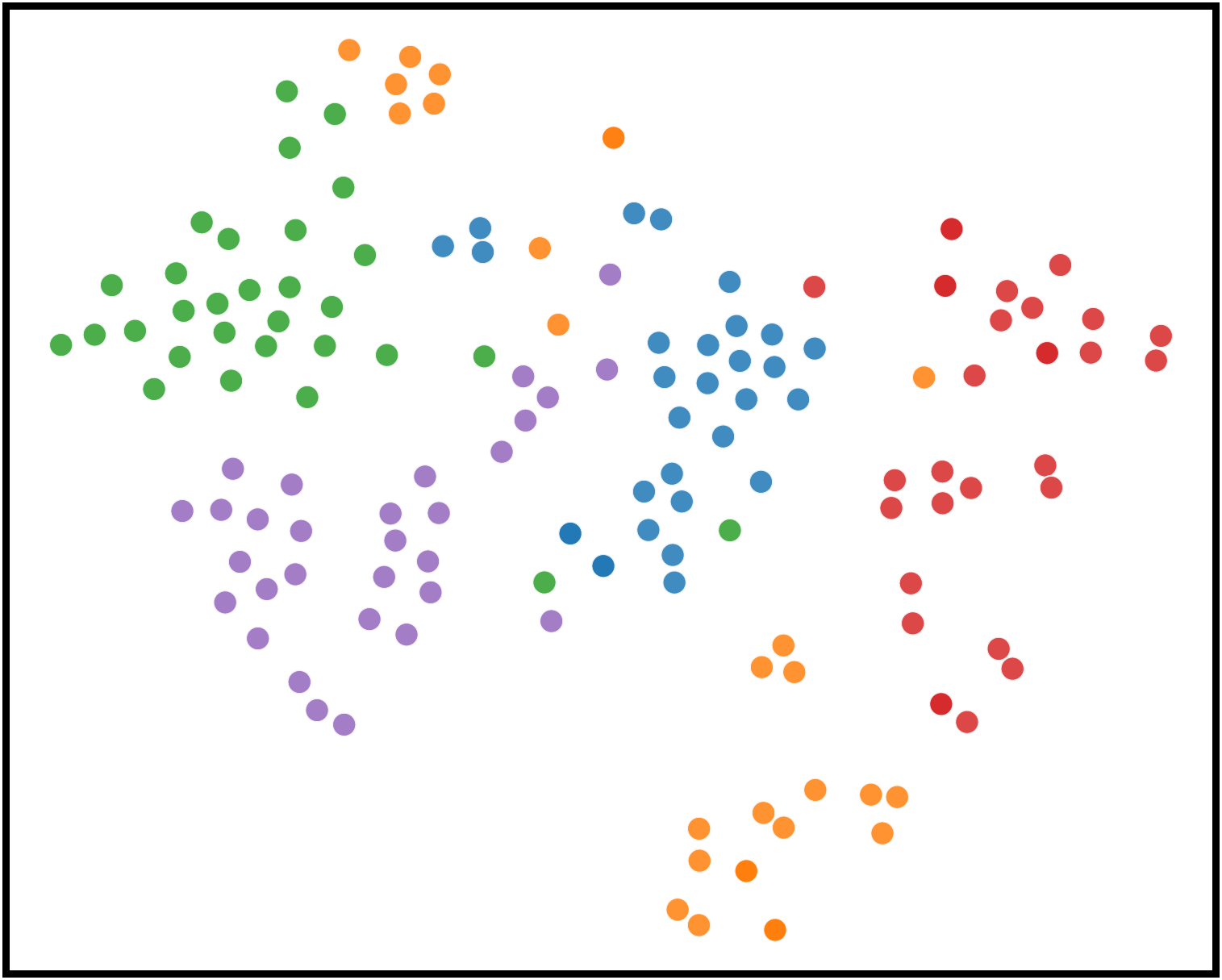}
    \caption{SE module}
    \label{fig:spatial}
\end{subfigure}
\hfill
\begin{subfigure}{0.19\textwidth}
    \centering
    \includegraphics[width=\linewidth]{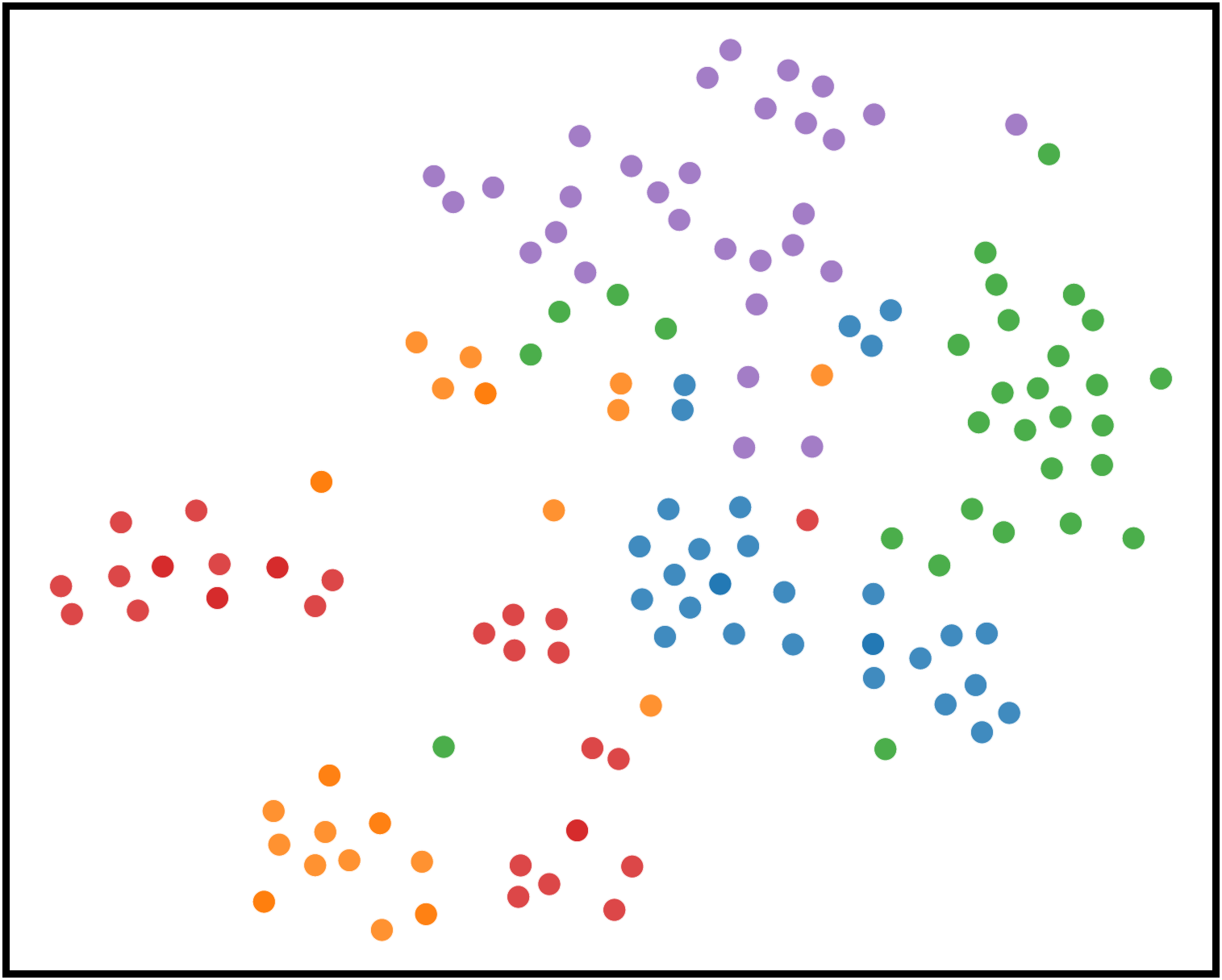}
    \caption{TMHA module}
    \label{fig:temporal}
\end{subfigure}
\hfill
\begin{subfigure}{0.19\textwidth}
    \centering
    \includegraphics[width=\linewidth]{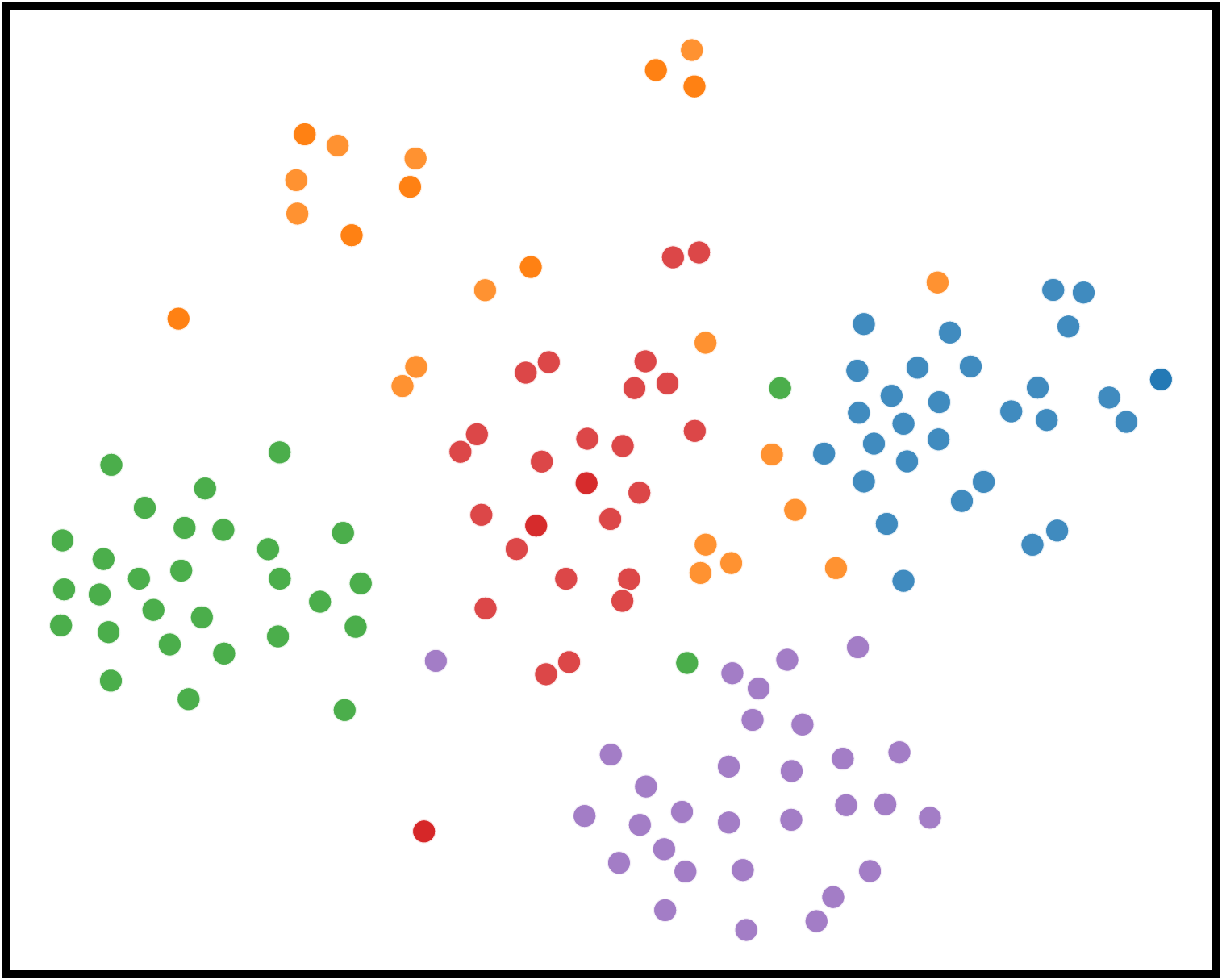}
    \caption{STFF module}
    \label{fig:fusion}
\end{subfigure}
\hfill
\begin{subfigure}{0.19\textwidth}
    \centering
    \includegraphics[width=\linewidth]{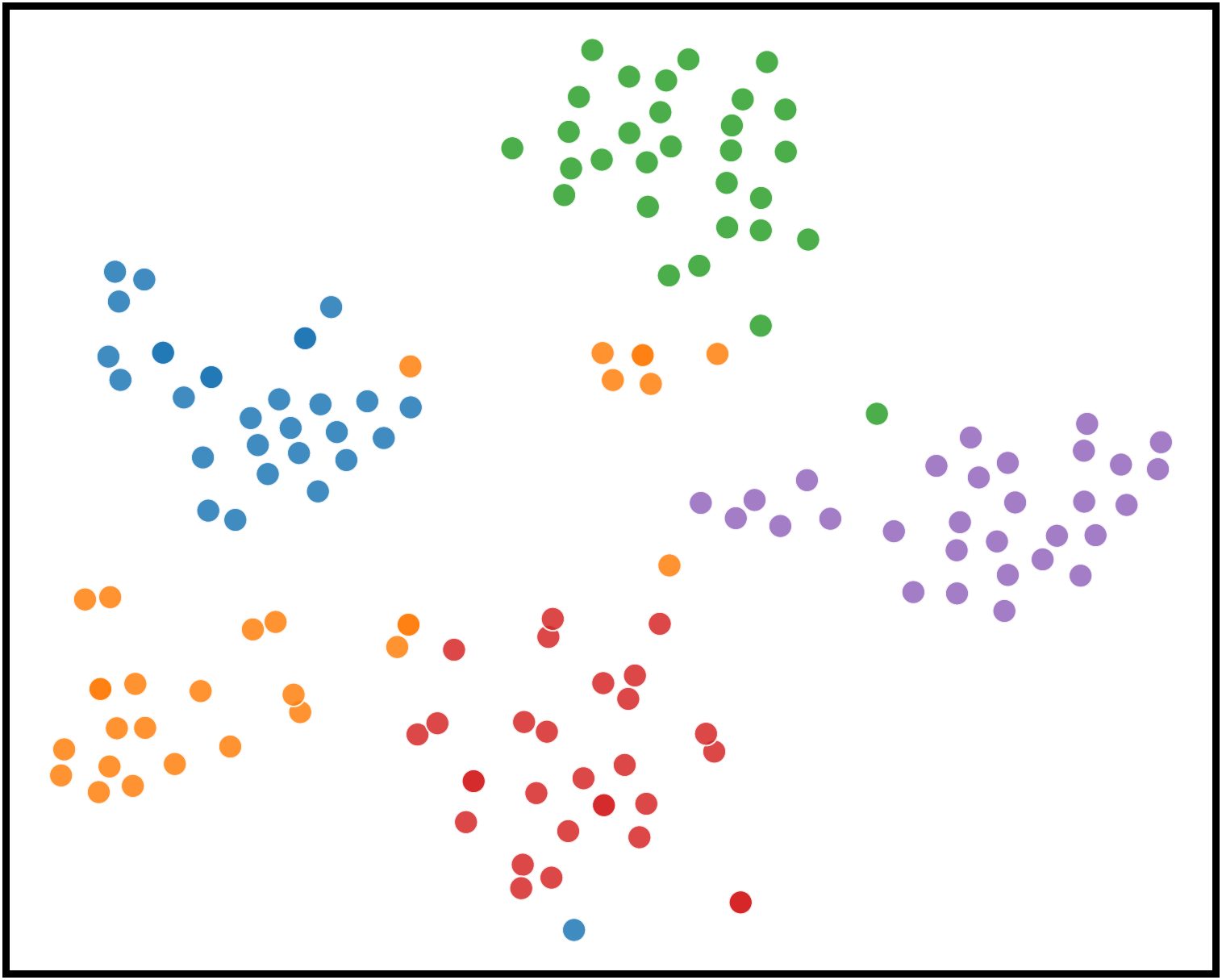}
    \caption{Ours}
    \label{fig:final}
\end{subfigure}
\caption{t-SNE visualization of feature representations learned by different modules on UCF101.}
\label{fig:tsne}
\end{figure*}

\subsubsection{\textbf{Hyperparameter Sensitivity Analysis}}
In Eqs. (\ref{eq:7}), (\ref{eq:11}) and (\ref{eq:17}), we introduce seven hyper‑parameters, i.e., \(\lambda\), \(\lambda_1\), \(\lambda_2\), \(\beta_1\), \(\beta_2\), \(\beta_3\), and \(\beta_4\), whose values affect the performance of the proposed method. To investigate their impact, we conduct parameter‑sensitivity analysis experiments in this section.

As shown in Fig. \ref{fig:lambda}, the proposed method is insensitive to the hyper‑parameter \(\lambda\). Although the recognition accuracy under both 1‑shot and 5‑shot settings exhibits slight fluctuations with the increase of \(\lambda\), the overall performance remains stable. As shown in Fig. \ref{fig:lambda12}, the accuracy fluctuates slightly within the search range of $[0.01,1]$, which demonstrates that our model is insensitive to the two hyper‑parameters. Specifically, the best performance is achieved at $\lambda_1=0.1$ and $\lambda_2=0.01$, while the lowest accuracy occurs under the extreme combination $(\lambda_1=0.01,\lambda_2=1.0)$. This indicates that these two hyper‑parameters exert limited influence on the final performance as long as extreme values are avoided.

As shown in Fig.~\ref{fig:param-beta}, we further investigate the influence of different loss weighting coefficients on the proposed method. Overall, the proposed method exhibits stable performance over a relatively wide range of parameter values. The highest recognition accuracy is consistently achieved when all loss weights are set to 0.1, indicating that a balanced optimization among center loss, alignment loss, contrastive loss, and dictionary loss is beneficial for learning discriminative video representations. When the loss weights are excessively small, the corresponding supervision signals become insufficient, whereas overly large weights tend to dominate the optimization process and slightly reduce the recognition accuracy.

\subsection{Visualization}
To further evaluate the discriminative capability of the learned video representations, we visualize the feature distributions of five action classes from the UCF101 dataset using t-SNE, as shown in Fig.~\ref{fig:tsne}. Fig.~\ref{fig:init} illustrates the initial feature distribution extracted by the MAE backbone. Although samples from several classes exhibit coarse grouping patterns, significant overlap and dispersion still exist among different categories, indicating that the backbone features alone are insufficient to distinguish visually similar actions under the few-shot setting.

Fig.~\ref{fig:spatial} presents the feature distribution after the SE module. Compared with the initial representations, samples belonging to the same class become noticeably more compact, while the inter-class boundaries become clearer. This improvement demonstrates that the SE module effectively captures video‑level global spatial information across frames and enhances the discriminability of spatial representations.

Fig.~\ref{fig:temporal} shows the representations learned by the TMHA module, the temporal features further improve the separation between action categories, especially for actions with similar appearance but different motion patterns. However, slight overlap between several neighboring classes can still be observed.

After heterogeneous spatial-temporal alignment and feature fusion, the learned representations shown in Fig.~\ref{fig:fusion} exhibit significantly improved intra-class compactness and inter-class separability, demonstrating that the proposed feature fusion strategy effectively integrates complementary spatial and temporal information into a unified representation.

Finally, Fig.~\ref{fig:final} presents the feature distribution produced by the complete model. Samples from the same class form compact clusters, while different classes are separated by larger margins with minimal overlap. This observation indicates that our proposed modules complement one another, and the hierarchical metric learning strategy helps produce more discriminative feature representations.

\section{Conclusion} \label{sec:5}
In this work, we have proposed a hierarchical metric learning framework named HML‑FSAR for few-shot action recognition. Firstly, we design a spatial‑enhanced module to capture cross‑frame global spatial representations. On this basis, we specially develop heterogeneous alignment and spatial‑temporal feature fusion modules to obtain more discriminative video‑level spatial‑temporal features, followed by a dictionary learning module to suppress feature noise. Secondly, built upon the learned feature representations, we propose a hierarchical metric learning strategy to impose multi‑stage complementary constraints for jointly optimizing feature compactness, heterogeneous spatial-temporal alignment capability, inter‑class discriminability and anti‑noise robustness, which explicitly guides feature learning via elaborate metric constraints. Experimental results demonstrate that our HML‑FSAR achieves competitive or state‑of‑the‑art performance on five standard benchmarks.

\section*{Acknowledgement}
The work was supported in part by the Shandong Provincial Natural Science Foundation under Grant Nos. ZR2025MS1004, ZR2025QC647; in part by the National Natural Science Foundation of China under Grants Nos. 62101213, 62471202.

\bibliographystyle{IEEEtran}
\bibliography{references}

\end{document}